\documentclass[journal]{IEEEtran}

\usepackage{xcolor,soul,framed} 
\usepackage{caption}
\colorlet{shadecolor}{yellow}
\usepackage[pdftex]{graphicx}
\graphicspath{{../pdf/}{../jpeg/}}
\DeclareGraphicsExtensions{.pdf,.jpeg,.png}
\usepackage{multirow}
\usepackage[cmex10]{amsmath}
\usepackage{array}
\usepackage{mdwmath}
\usepackage{mdwtab}
\usepackage{eqparbox}
\usepackage{url}
\usepackage{hyperref}
\usepackage{makecell}
\usepackage{amsmath}
\usepackage{flexisym}
\usepackage{amssymb}
\usepackage{acronym}
\usepackage{algorithmic}
\usepackage[]{footmisc}
\usepackage{mdwmath}
\usepackage{mdwtab}
\usepackage{fixltx2e}
\usepackage{dblfloatfix}
\usepackage{bm}
\usepackage{hyperref}
\usepackage{caption}
\usepackage{url}
\usepackage{eqparbox}
\newcommand{\revision}[1]{\textcolor{black}{#1}}

\usepackage{tikz}
\usetikzlibrary{decorations.text}

\newcommand\copyrighttext{%
  \small \textcopyright 2023 IEEE. Personal use of this material is permitted. Permission from IEEE must be obtained for all other uses, in any current or future media, including reprinting/republishing this material for advertising or promotional purposes, creating new collective works, for resale or redistribution to servers or lists, or reuse of any copyrighted component of this work in other works. DOI: \href{https://doi.org/10.1109/TCSVT.2023.3255243}{10.1109/TCSVT.2023.3255243}}
\newcommand\copyrightnotice{%
\begin{tikzpicture}[remember picture,overlay]
\node[anchor=south, yshift=2pt] at (current page.south) {\fbox{\parbox{\dimexpr\textwidth-\fboxsep-\fboxrule\relax}{\copyrighttext}}};
\end{tikzpicture}%
}

\begin{document}
\title{SARGAN: Spatial Attention-based Residuals for Facial Expression Manipulation}
\author{\IEEEauthorblockN{Arbish Akram and Nazar Khan} \\
\IEEEauthorblockA{Department of Computer Science, University of the Punjab, Pakistan}}

\markboth{IEEE TRANSACTIONS ON CIRCUITS AND SYSTEMS FOR VIDEO TECHNOLOGY, VOL. ,NO. ,MARCH ~2023
}{}

\maketitle

\begin{abstract}
Encoder-decoder based architecture has been widely used in the generator of generative adversarial networks for facial manipulation. However, we observe that the current architecture fails to recover the input image color, rich facial details such as skin color or texture and introduces artifacts as well. In this paper, we present a novel method named SARGAN that addresses the above-mentioned limitations from three perspectives. First, we employed spatial attention-based residual block instead of vanilla residual blocks to properly capture the expression-related features to be changed while keeping the other features unchanged. Second, we exploited a symmetric encoder-decoder network to attend facial features at multiple scales. 
Third, we proposed to train the complete network with a residual connection which relieves the generator of pressure to generate the input face image thereby producing the desired expression by directly feeding the input image towards the end of the generator. Both qualitative and quantitative experimental results show that our proposed model performs significantly better than state-of-the-art methods. In addition, existing models require much larger datasets for training but their performance degrades on out-of-distribution images.
In contrast, SARGAN can be trained on smaller facial expressions datasets, which generalizes well on out-of-distribution images including human photographs, portraits, avatars and statues. 
\end{abstract}


\begin{figure}[t]
    \centering
    \includegraphics[width=.9\linewidth]{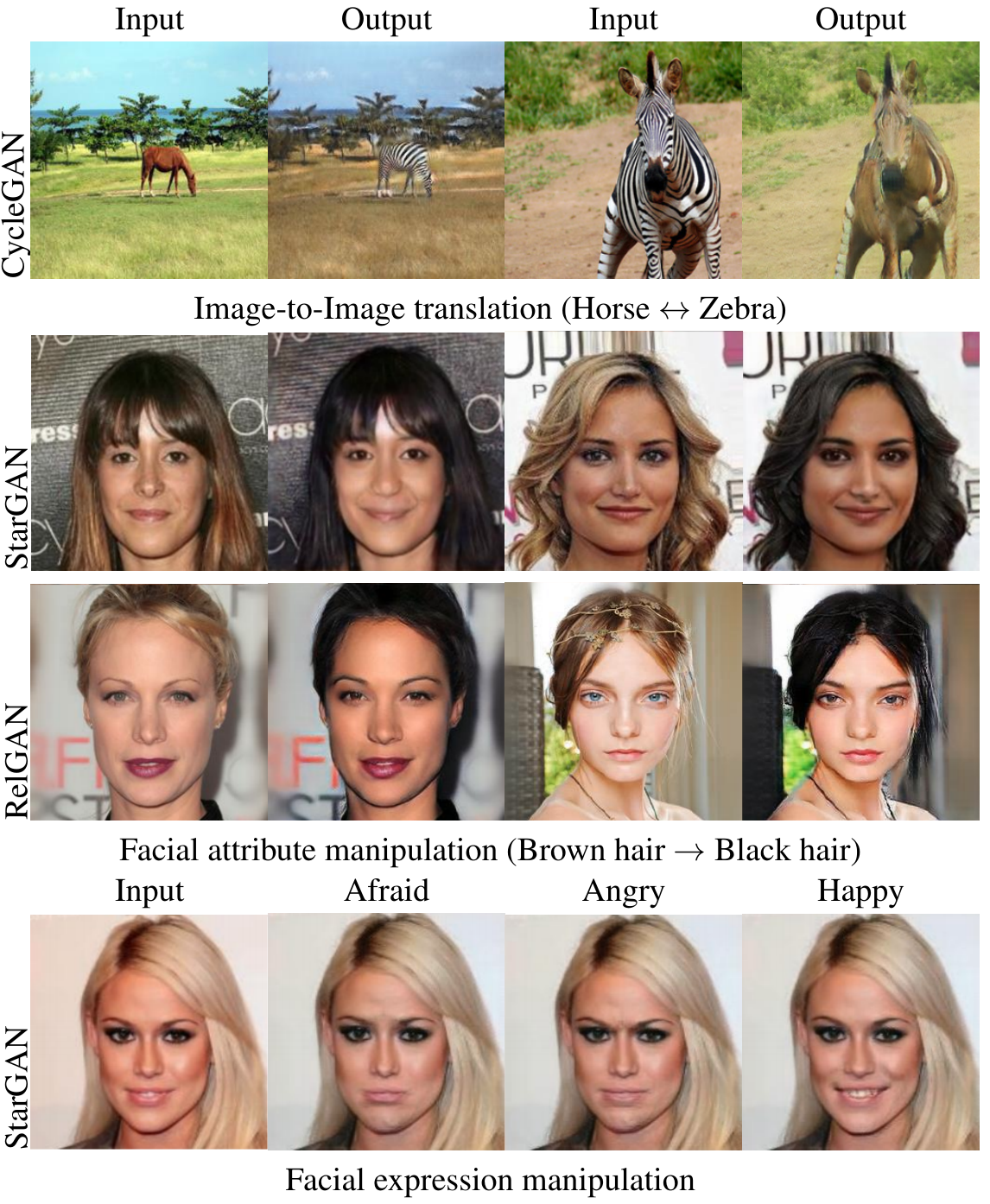}
    \caption{Weaknesses of state-of-the-art GAN-based models. These results show that though CycleGAN transforms horse into zebra and zebra into horse, it fails in preserving the background scene in its true colors. StarGAN and RelGAN successfully introduce the desired attribute but they also change skin tone and eye color. Similarly, StarGAN introduces realistic expressions but remains unable to preserve rich facial details such as skin tone and background color.}    
    \label{fig:weaknesses-models}
\end{figure}


\section{Introduction}
\IEEEPARstart{F}{acial} expression manipulation is the process of manipulating the expression-related features of an input facial image while keeping the non-expression-related features unchanged.
Recently, the field of image synthesis has achieved remarkable progress subsequent to the emergence of generative adversarial network (GAN) \cite{goodfellow-2014} and its conditional variant (cGAN) \cite{mirza2014conditional}. GANs have proven successful for face-related tasks including facial attribute editing \cite{liu2015faceattributes, shen-2016, he2019attgan, gao2021high} and face super-resolution \cite{chen2018fsrnet,  hu2020face, hu2021face, chen2020learning}. A substantial amount of work has also been done in the area of facial expression synthesis with the prevalence of GANs such as StarGAN \cite{choi-2017}, GANimation \cite{pumarola2018ganimation}, Cascade-EF GAN \cite{wu2020cascade}, DAI2I \cite{chen2020domain} and LGP-GAN \cite{xia2021local}. However, most of these methods including StarGAN, GANimation, Cascade-EF GAN and DAI2I need large training datasets. LGP-GAN \cite{xia2021local}, that can be trained on smaller expression synthesis datasets, fails to perform well on out-of-distribution images. 

\copyrightnotice

\begin{figure}[t]
    \centering
    \includegraphics[width=.95\linewidth]{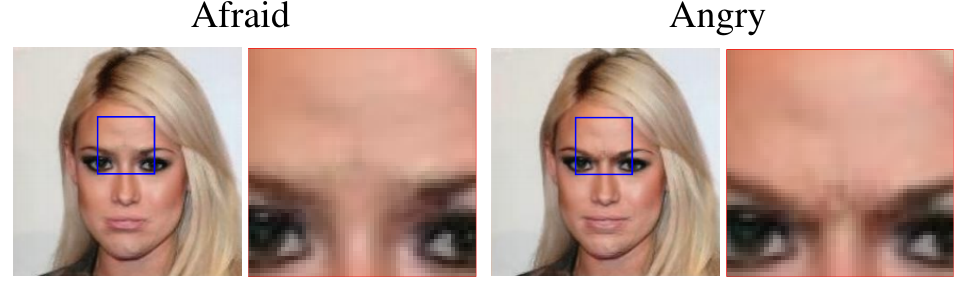}
    \caption{StarGAN tends to generate artifacts around expression-rich regions.}
    \label{fig:weakness_2}
\end{figure}

We take three state-of-the-art models CycleGAN \cite{zhu-2017}, StarGAN \cite{choi-2017} and RelGAN \cite{wu2019relgan} that share a common architecture in their generator. CycleGAN learns the mapping between two domains while StarGAN and RelGAN learn the mappings among multiple domains, in an unsupervised fashion by employing the cycle consistency loss. Fig. \ref{fig:weaknesses-models} demonstrates that CycleGAN successfully transforms horse into zebra and zebra into horse. However, it fails to preserve the background scene in its true colors in the generated images. Similarly, when we feed face images to StarGAN \cite{choi-2017} and RelGAN \cite{wu2019relgan} and expect them to only change hair color from brown to black, they also change eye color, skin tone and the background (see Fig. \ref{fig:weaknesses-models}, Rows 2 and 3). It also demonstrates that StarGAN is unable to preserve the facial and overall color details of the input image as well as introduces artifacts (see Fig. \ref{fig:weakness_2}). While such modifications are acceptable for image-to-image translation problems, they significantly affect the results of facial manipulation.

The weaknesses of existing image-to-image translation models can be summarised as follows:

\begin{enumerate}
    \item Existing state-of-the-art facial manipulation models are unable to preserve the facial and overall color details of the input images.
    \item State-of-the-art FES methods tend to generate artifacts around expression-intensive regions such as eyes, nose and mouth.
\end{enumerate}

The generator of these models consists of three modules: encoding layers, residual blocks \cite{he2016deep} and decoding layers. Encoding layers enable the network to encode salient facial details, whereas decoding layers recover high-level facial details and generate an output image that is most similar to the desired output image. The above-mentioned weaknesses of this architecture indicate that decoding layers are unlikely to recover the true facial colors of the input images because too much information is lost during convolution. Therefore, advancements are needed to address the drawbacks of the current encoder-decoder architecture.

We propose to directly link the input image to an output image to force the network to learn only the residual of the problem. By doing this, the generator network just encodes and decodes information related to the expression rather than simultaneously learning information relating to the expression and the input image facial and color details. Since the problem has been simplified now, we found that it can be solved now with only \textbf{one} residual block as opposed to six in competing models.  

\begin{figure}[t]
    \centering
    \includegraphics[width=.95\linewidth]{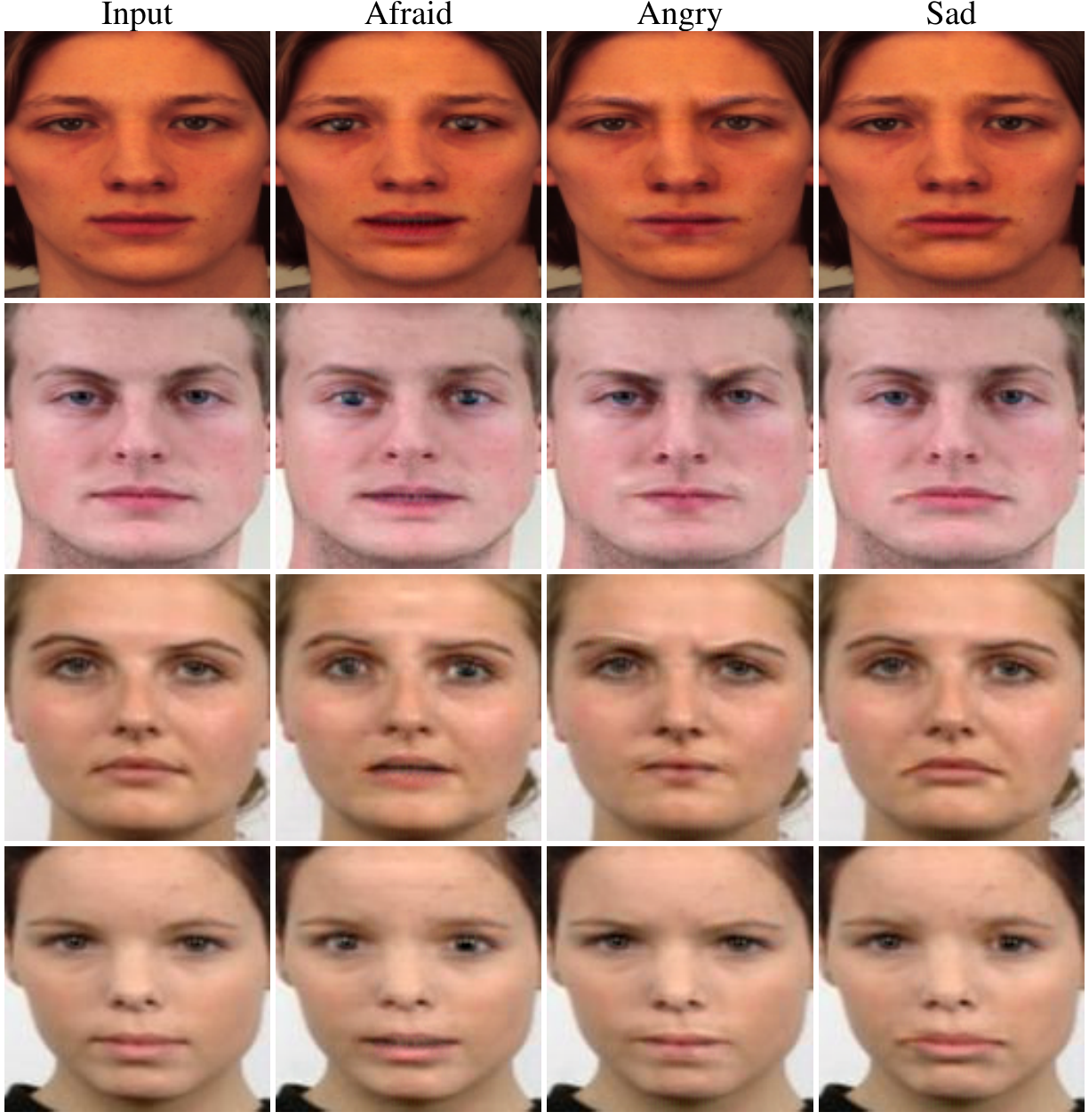}
   \caption{Results of facial expression manipulation using our proposed SARGAN model on KDEF \cite{lundqvist-1998} (row 1), CFEE \cite{du-2014} (row 2) and RaFD \cite{langner-2010} (rows 3 and 4) testing images. SARGAN effectively transforms expressions while preserving facial details including skin tone and eye color.}
    \label{fig:indataset-results}
\end{figure}

Further, we propose to exploit a spatial attention-based residual block (SARB) rather than vanilla residual block for expression manipulation. SARB focuses on expression-related regions only instead of the complete face. We use a symmetric encoder-decoder network to compute an attention map in order to decide which spatial regions should be emphasized for expression manipulation.  This symmetric encoder-decoder network symmetrically connects the encoding and decoding layers using skip connections, allowing the training procedure to converge substantially faster and aiding in the consolidation of facial features across many scales while preserving spatial information. The results of our proposed framework SARGAN are presented in Fig. \ref{fig:indataset-results}. These results demonstrate that the proposed method i) correctly induced the desired expressions, ii) did not introduce artifacts, and iii) preserved rich facial details. Our contributions can be summarised as follows:
\begin{enumerate}
    \item Proposed to train the complete network with skip connection which forces the generator to learn only expression-related details.
    \item Incorporated spatial-attention based residual block to properly capture expression-related features required to be changed while keeping the other features unchanged.
    \item Used a symmetric encoder-decoder network followed by a convolutional layer to build an attention map to attend high and low level expression-related features.
    \item Demonstrated that SARGAN has good generalization capability over out-of-distribution unconstrained images covering arbitrary celebrities, portraits, statues, avatars as well as images containing self-occlusion by hand or other external objects.
    \item Exhibited that proposed architecture is also useful for facial attribute manipulation.
\end{enumerate}


The rest of the paper is structured as follows. Section \ref{related_work} reviews the state-of-the-art related to facial expression manipulation. In Section \ref{proposed_method}, we provide the details of our proposed SARGAN. Qualitative as well as quantitative experimental results of the proposed method on in- and out-of-distribution images are presented in Section \ref{exp_and_results} and the paper is concluded in Section \ref{conclusion}.

\section{Related Work}
\label{related_work}
\textbf{Generative Adversarial Network}: In recent years, the entire  research in image synthesis has seen considerable advancement after the emergence of Generative Adversarial Network (GANs) \cite{goodfellow-2014} and its conditional variant \cite{mirza2014conditional}. GANs have demonstrated their capabilities in various computer vision tasks including image synthesis \cite{brock2018large, yuan2019bridge, karras2017progressive, zhang-2017stackgan}, image-to-image translation \cite{zhang2019image, pang2018visual, isola-2016, zhu-2017}, image in-painting \cite{yeh2017semantic, yu2018generative, yan2018shift} and face aging \cite{sun2020facial, liu2017face, shi2020can}. 
A typical GAN-based model simultaneously trains a generator and a discriminator in an adversarial manner. During this adversarial learning, the generator tries to produce the photorealistic images while the discriminator learns to identify the differences between real and synthetic images. Conditional variants of GAN \cite{mirza2014conditional} include conditional information in the form of text, images, labels etc to control the generated images. Facial attribute manipulation and facial expression manipulation are well-studied topics these days.

\revision{\textbf{Attention Mechanisms}:
Attention in images refers to the adaptive selection of specific parts or features of an image while ignoring the other ones. In the literature, three types of attention mechanisms are frequently used for images: i) channel attention \cite{hu2018squeeze, zhang2018image}, ii) temporal/sequential attention \cite{xu2015show, anjum2020urdu_ohtr}, 
and iii) spatial attention \cite{chen2020learning, chen2017sca}. Channel attention dynamically weighs the different channels of convolutional feature maps. 
Temporal/sequential attention weighs either temporal or sequential portions of its input. Spatial attention focuses on important 2D regions of the input while ignoring others. In this work, we utilize spatial attention in the residual block to focus on expression-related features of the input image to enhance the visual quality of the generated facial expressions.}

\textbf{Facial Attribute Manipulation}: Most facial manipulation methods formulate this problem as unpaired image-to-image translation. Authors, in \cite{perarnau-2016, liu-2017, larsen-2015, shen-2016, zhang2021joint}, have tackled facial attribute manipulation by modifying output image attributes such as gender swapping, changing hair color, adding and removing glasses. AttGAN \cite{he2019attgan} trains an encoder-decoder architecture to translate facial attributes based on the target attributes. STGAN \cite{liu2019stgan} symmetrically links encoding and decoding layers with selective transfer units for facial attribute manipulation.
RelGAN \cite{wu2019relgan}, a multi-domain image-to-image translation framework, uses relative-attribute information and matching-aware discriminator to transfer facial attributes. HifaFace \cite{gao2021high} framework revisits cycle-consistency loss, and uses wavelet-based generator and high-frequency discriminator for high-fidelity facial attribute manipulation.

\textbf{Facial Expression Manipulation}: Our work pertains to facial expression manipulation models. Facial expression manipulation has seen tremendous success with the advent of GAN-based frameworks. In this regard, Song et al. \cite{song2018geometry} proposed a Geometry-Guided GAN (G2-GAN) that employs facial landmarks as a controllable condition to control the intensity of the generated facial expressions. However, target face geometrical information is essential for expression manipulation in G2-GAN. The authors in \cite{ding2018exprgan}, introduced an Expression GAN (ExprGAN) to generate expressions in discrete and continuous domains. StarGAN \cite{choi-2017}, a multi-domain image-to-image translation model, uses a single generator to perform facial expression manipulation among seven discrete emotions. GANimation \cite{pumarola2018ganimation}, incorporates attention mechanism to change expression-related pixels in input image. It also adopts Action Units (AUs) as emotion labels and can generate expressions in a continuous domain. However, both these methods tend to generate artifacts on out-of-distribution images. TFEM \cite{ling2020toward} improves the results of GANimation by using relative AUs as conditional input. In addition, they also introduce multi-resolution feature fusion mechanism in U-Net architecture. SMIT \cite{romero2019smit}, a stochastic multi-label image-to-image translation model generates many outputs for a single input image by employing stochastic noise-driven manipulation. In \cite{wu2020cascade}, the authors proposed to focus on the local regions of face and employed progressive editing strategy to generate realistic and sharp facial expressions. 
GANmut \cite{d2021ganmut} introduced the idea to let the model learn the emotion label space instead of using hand-crafted labels. The authors in \cite{xia2021local} proposed a two-step model, LGP-GAN, to synthesize realistic discrete expressions. However, they also mentioned that LGP-GAN introduces realistic expression on the lab-collected datasets and does not perform well on out-of-dataset images. In contrast, our proposed method has good generalization capability despite being trained on lab-collected images. A domain adaptive image-to-image translation framework has been presented by Chen et al. \cite{chen2020domain}
to induce expression on different style images. Zhang et al. \cite{zhang2021joint} designed a deep GAN-based network, a facial expression transfer method, that takes two images as input and learns to solve expression synthesis transfer and recognition problems simultaneously. On the contrary, we attempt to tackle the problem of facial expression synthesis instead of facial expression transfer. Peng et al. \cite{peng2021unified} proposed a unified framework for high fidelity face swap and face reenactment by decomposing a face image into 3D parameters of pose, shape and expression.  These 3D parameters were then recombined to produce the results of the face swap and face reenactment. Instead of splitting a face image into 3D parameters, we solve the expression translation problem by adapting it to a GAN-based framework. 

Recently, regression-based methods have been also proposed for facial expression manipulation. For example, MR \cite{khan2020masked} employs a constrained version of regression that considers local regions of input images to make output units. In \cite{akram2021-pixel_fes}, the authors proposed that only one fixed input unit should be observed to produce the output of one unit. However, these regression-based methods require carefully-aligned images and post-processing refinement steps to recover sharp details of the face image. 

\begin{figure*}[t]
    \centering
    \includegraphics[width=.9\linewidth]{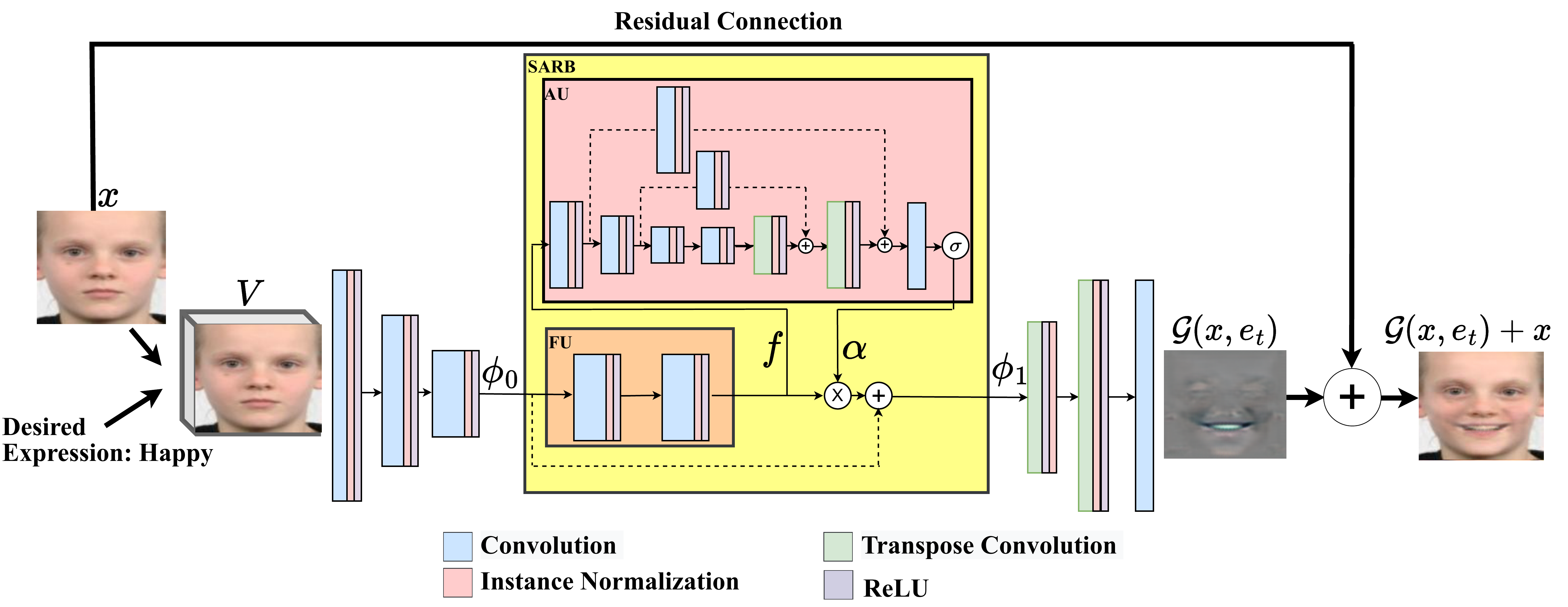}
    \caption{Illustration of our proposed generator $\mathcal{G}$. SARGAN takes a face image and the desired expression vector as input and produces an output image with the desired expression.}
    \label{fig:sargan_arch}
\end{figure*}

\section{Proposed Method}
\label{proposed_method}
Facial expression manipulation (FEM) models (e.g StarGAN \cite{choi-2017}) take a face image and the desired expression as input and translate it into an output image with the result that the output face image looks photo-realistic and has the desired expression. StarGAN constitutes an encoder-decoder architecture in its generator which is adapted from CycleGAN \cite{zhu-2017}. CycleGAN has shown impressive results on image-to-image translation problems including horse to zebra transfiguration, season and painting style transfer. However, we observe an intriguing phenomenon as follows. When this architecture has been used for face-to-face translation problems, it introduces artifacts and skin tone color degradation problem as shown in Fig. \ref{fig:weaknesses-models}. It shows that this architecture is not capable of preserving the facial details in the synthesized images. These problems cannot be tolerated in face-related task. 

We address the aforementioned problems from two perspectives.
\begin{enumerate}
    \item We incorporate attention within the residual block so that the model learns to focus on expression-related regions instead of the whole face.
    \item We feed the input image directly into the output of the generator via a residual connection. This spares the generator from trying to retain input-related details such as identity, facial details, and overall color distribution. As a result the generator learns expression-related features only.
\end{enumerate}

Fig. \ref{fig:sargan_arch} shows the overall architecture of our proposed framework's  generator $\mathcal{G}$. 
Given an input image $\bm{x} \in \mathbb{R}^{H \times W \times 3}$ and the desired expression vector $\bm{e}_t$ as a 1-hot vector of $E$ expressions, we form our input volume $V\in \mathbb{R}^{H \times W \times (3 + E)}$ by concatenating the image with a 1-hot channel representation of the desired expression. Our proposed network's generator consists of one convolution that maps volume $V$ into features, two strided convolution having stride size two for downsampling, one spatial attention-based residual block, two transposed convolutions with stride size two for upsampling and one convolution that maps features to an RGB image. 

\subsection{Spatial Attention-based Residual Block}
Spatial attention-based residual block (SARB) employs attention mechanism to change only expression-related features. SARB can be divided into two units: Feature unit and Attention unit.

\textbf{Feature Unit (FU)} 
produces a feature volume $\bm{f}$ by passing the intermediate volume $\bm{\phi}_0$ through two convolutional layers, each followed by instance normalization and ReLU layers.  

\textbf{Attention Unit (AU)} determines the importance of different facial features in $\bm{f}$.
A spatial attention map $\bm\alpha$ can be produced by utilizing a symmetric encoder-decoder network (SEDN) followed by a convolutional layer. The encoder of SEDN takes the feature volume $\bm{f}$ and extracts features at multiple scales. 
To obtain the output of the same size as the feature volume $\bm{f}$, the decoder tries to reconstruct the feature volume $\bm{f}$ from low-level encoding by using transposed convolutions. Decoding layers cannot reconstruct the details of the original feature volume $\bm{f}$ because feature details are already lost in the encoding process due to subsampling. Therefore, we link the encoding layers to their corresponding decoding layers via residual connections. These symmetric residual connections help to recover better and more meaningful facial features. In this way, SEDN consolidates facial features across multiple scales while preserving spatial information. \revision{This type of network has performed well for tasks such as face alignment \cite{bulat2017far}, face parsing \cite{chen2018fsrnet} and face super-resolution \cite{chen2020learning}. Therefore, we exploit such a configuration for facial expression manipulation.}
The details of the AU are provided in Fig. \ref{fig:sargan_arch}. Similar ideas have been proposed in \cite{lin2017feature, newell2016stacked}. 

To concentrate more on informative features and suppress uninformative features, a gating mechanism over the output $\bm{f}$ of the feature unit produces the encoding $\bm{\phi}_1$ of the input image $\bm{x}$ conditioned on the desired expression $\bm{e}$ as
\begin{align}
    \bm{\phi}_1 &= \bm{\phi}_0 + \bm\alpha \otimes \bm{f},
    \label{eq:sarb-output}
\end{align}
where $\otimes$ represents element-wise multiplication. We feed the output $\bm{\phi}_1$ of SARB to transposed convolutional layers and get the output image $\mathcal{G}(\bm{x}, \bm{e}_t)$.

\subsection{Residual Connection}
Residual connections have achieved remarkable success in image-to-image translation \cite{zhu-2017, choi-2017, pumarola2018ganimation} and style transfer tasks \cite{johnson-2016perceptual}. Since input and output images share the same subject, residual connections preserve the exact details of the input image.    
We propose to train the generator of SARGAN with a residual connection. This obviates the need for the generator to produce input image with the desired expression by enforcing only expression-related details. Residual connection is usually applied at the feature level, but here we used residual connection at the RGB image level. The output image can be formulated as
\begin{align}
    \bm{y} &= \mathcal{G}(\bm{x}, \bm{e}_t) + \bm{x}.
\end{align}

In this way, instead of directly learning the mapping  $\mathcal{G}(\bm{x}, \bm{e}_t) \rightarrow \bm{y}$, we will learn only the residual of the problem $\mathcal{G}(\bm{x}, \bm{e}_t) \rightarrow \bm{y} - \bm{x}$.

\subsection{Objective Function}
We trained SARGAN by minimizing the following three loss functions. 

\textbf{Adversarial Loss}:
To learn the parameters of the generator $\mathcal{G}$ and discriminator $\mathcal{D}$, we utilize Wasserstein GAN loss with gradient penalty \cite{gulrajani2017improved} defined as
\begin{align}
    \mathcal{L}_{adv}=& \mathbb{E} 
   \left[ \mathcal{D}_{r/s}(\bm{x})\right] - \mathbb{E}\left[\mathcal{D}_{r/s}(\mathcal{G}(\bm{x}, \bm{e}_t)+\bm{x})\right] \\ \nonumber
   & - \lambda_{gp} \mathbb{E} \left[ (\Vert \nabla_{\bm{\hat{x}}} \mathcal{D}_{r/s}(\bm{\hat{x}}) \Vert_2 - 1)^2
    \right],
\end{align}
where $\lambda_{gp}$= 10 is a penalty coefficient. 

\textbf{Cycle Consistency Loss}:
Let $\bm{\hat{x}}$ represent the reconstruction of the input image $\bm{x}$ generated from the synthesized expression $\bm{y}$ as
\begin{align}
    \bm{\hat{x}} =& \mathcal{G}(\mathcal{G}(\bm{x}, \bm{e}_t)+\bm{x},\bm{e}_o)+\bm{x}.
\end{align}

By minimizing adversarial loss, the generator is forced to generate photo-realistic facial expressions. However, it does not guarantee that the synthesized face and the input face correspond to the same person. To constrain this property, we adopt cycle consistency loss \cite{zhu-2017} to minimize the difference between input image and its  reconstruction.
\begin{align}
    \mathcal{L}_{cyc}=& \mathbb{E}
    \left[ \Vert \bm{x} - \bm{\hat{x}} \Vert_1 \right].
\end{align}

\textbf{Expression Classification Loss}:
A multiclass cross-entropy loss between the original expression $\bm{e}_o$ and the generated expression $\bm{e}_o\textprime$ of the input image $\bm{x}$ is defined as
\begin{align}
    \mathcal{L}_{ec}^{real} =& \mathbb{E} 
    \left[ - \log \mathcal{D}_{ec} (\bm{e}_o|\bm{x}, \bm{e}_o\textprime)
    \right].
\end{align}

While the loss between the target expression $\bm{e}_t$ and classified expression $\bm{e}_t\textprime$ of the synthesized image $\mathcal{G}(\bm{x}, \bm{e}_t)+\bm{x}$ is computed as
\begin{align}
    \mathcal{L}_{ec}^{syn} =& \mathbb{E} \left[ - \log \mathcal{D}_{ec} (\bm{e}_t|\mathcal{G}(\bm{x}, \bm{e}_t)+\bm{x}, \bm{e}_t\textprime)
    \right].
\end{align}

\textbf{Full Loss}:
The full loss function for $\mathcal{D}$ and $\mathcal{G}$ can be written as
\begin{align}
    \mathcal{L}_{\mathcal{G}} &= \mathcal{L}_{adv} + \lambda_{ec} \mathcal{L}_{ec}^{syn} +  \lambda_{cyc} \mathcal{L}_{cyc}, \\ \nonumber
    \mathcal{L}_{\mathcal{D}} &= - \mathcal{L}_{adv} + \lambda_{ec} \mathcal{L}_{ec}^{real} .
    \label{eq:full_loss}
\end{align}
 
The ability of $\mathcal{D}$ to discriminate between real and synthesized images and classify the expression of the discriminator's input image are improved by minimizing $\mathcal{L}_{D}$. While the generator $\mathcal{G}$ is urged to produce synthesized images that are hard to distinguish from real images, have the desired expression and preserve the characteristics of the input images.

\section{Experiments and Results}
\label{exp_and_results}

\subsection{Dataset}
\begin{figure}[t]
    \centering
    \includegraphics[width=\linewidth]{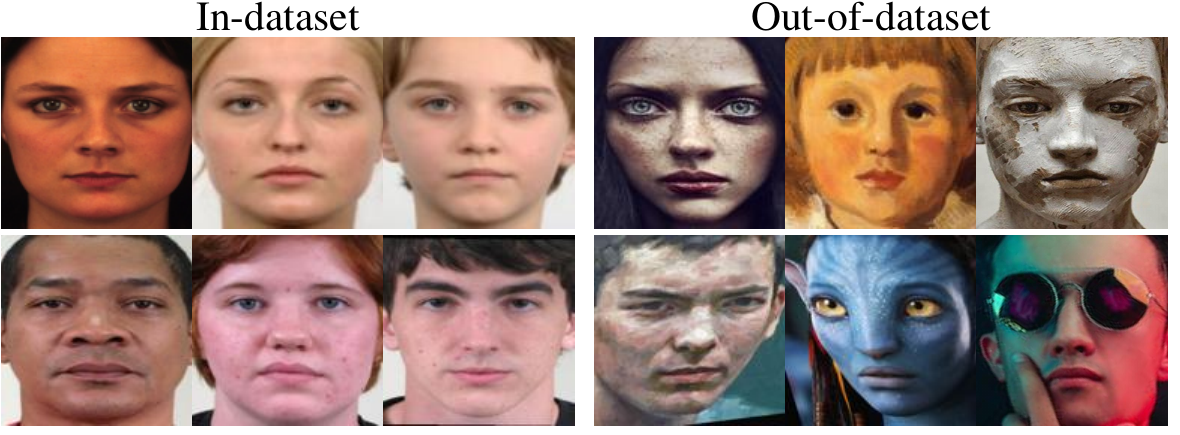}
    \caption{Samples images from in- and out-of-dataset.}
    \label{fig:dataset_fig}
\end{figure}

In order to assess the performance of our proposed framework, we conduct experiments on three publically available facial expression databases: CFEE \cite{du-2014}, KDEF \cite{lundqvist-1998} and RaFD  \cite{langner-2010}. These datasets include seven universal expressions including afraid, angry, disgusted, neutral, happy, sad and surprised. We obtained $2,569$ images for all facial expressions. The images are center-cropped and resized to $128 \times 128$ for all experiments. We randomly select $90\%$ images for training and $10\%$ for testing.  We performed color-based data augmentation via ReHistoGAN \cite{afifi2021histogan}. To assess the generalization capability of SARGAN on out-of-distribution images, we downloaded the images of arbitrary celebrities, portraits, avatars, sculptures and statues from the Internet. Fig. \ref{fig:dataset_fig} shows sample images from in- and out-of-dataset images.

\subsection{Baselines}
We compare our proposed method with five state-of-the-art facial expression manipulation methods including StarGAN \cite{choi-2017}, STGAN \cite{liu2019stgan}, GANimation \cite{pumarola2018ganimation}, TFEM \cite{ling2020toward} and DAI2I \cite{chen2020domain}. We trained StarGAN, STGAN and TFEM on the same dataset that was used to trained SARGAN. TFEM uses AUs as emotion labels, we obtain these labels by \cite{baltrusaitis2018openface}. To obtain results for DAI2I and GANimation, we used their pre-trained models. 

\begin{figure*}[ht]
    \centering
    \includegraphics[width=.9\linewidth]{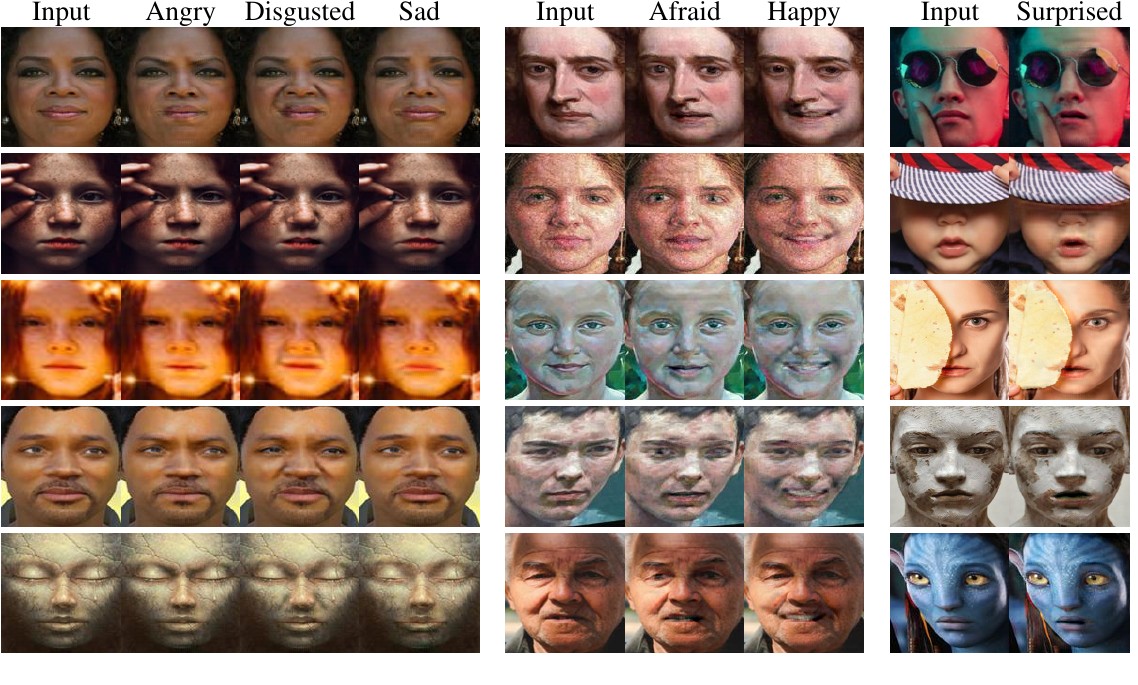}
    \caption{Results of SARGAN model on out-of-distribution unconstrained images covering arbitrary celebrities, portraits, statues, avatars and images containing self occlusion by hand or other external objects. These results show that SARGAN introduces desired expression on a variety of images while preserving identities, facial details and overall color distribution.}
    \label{fig:outofdataset-results}
\end{figure*}

\subsection{Implementation and Hyper-parameters details}
We implement our proposed method in Pytorch framework. The network is trained using Adam optimizer with learning rate $0.0001$, $\beta_1=0.5$, $\beta_2=0.999$ and batch size of $8$ for $400$ epochs. During training, the generator is updated after updating the discriminator five times as in \cite{gulrajani2017improved}. For the discriminator, we adopted the architecture introduced in \cite{choi-2017} for facial expression manipulation. This discriminator has two branches: i) $\mathcal{D}_{r/s}$ discriminates between real and synthesized images, ii) $\mathcal{D}_{ec}$ estimates probabilities representing the expression of the discriminator's input image. 
We used a grid search method between $0$ and $100$ to find the best values for $\lambda_{ec}$ and $\lambda_{cyc}$ in Eq. 8
. The cross-validated weight coefficients used are $\lambda_{ec}=1$ and $\lambda_{cyc}=10$.

\begin{figure}[t]
    \centering
    \includegraphics[width=.95\linewidth]{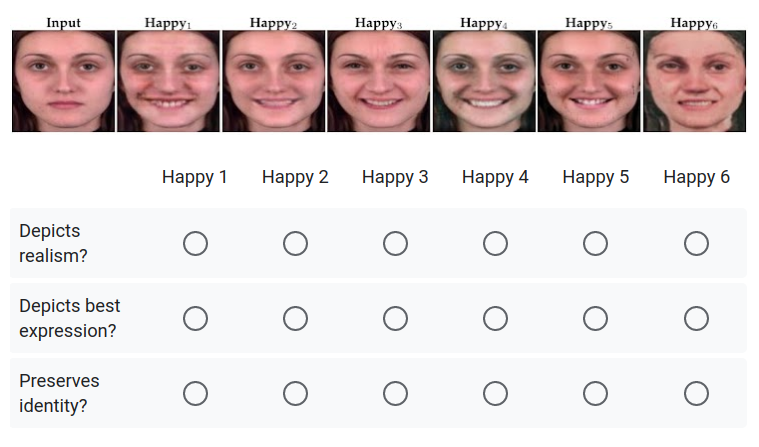}
    \caption{Example image from user study.}
    \label{fig:sample_user_study}
\end{figure}

\begin{figure}[t]
    \centering
    \includegraphics[width=\linewidth]{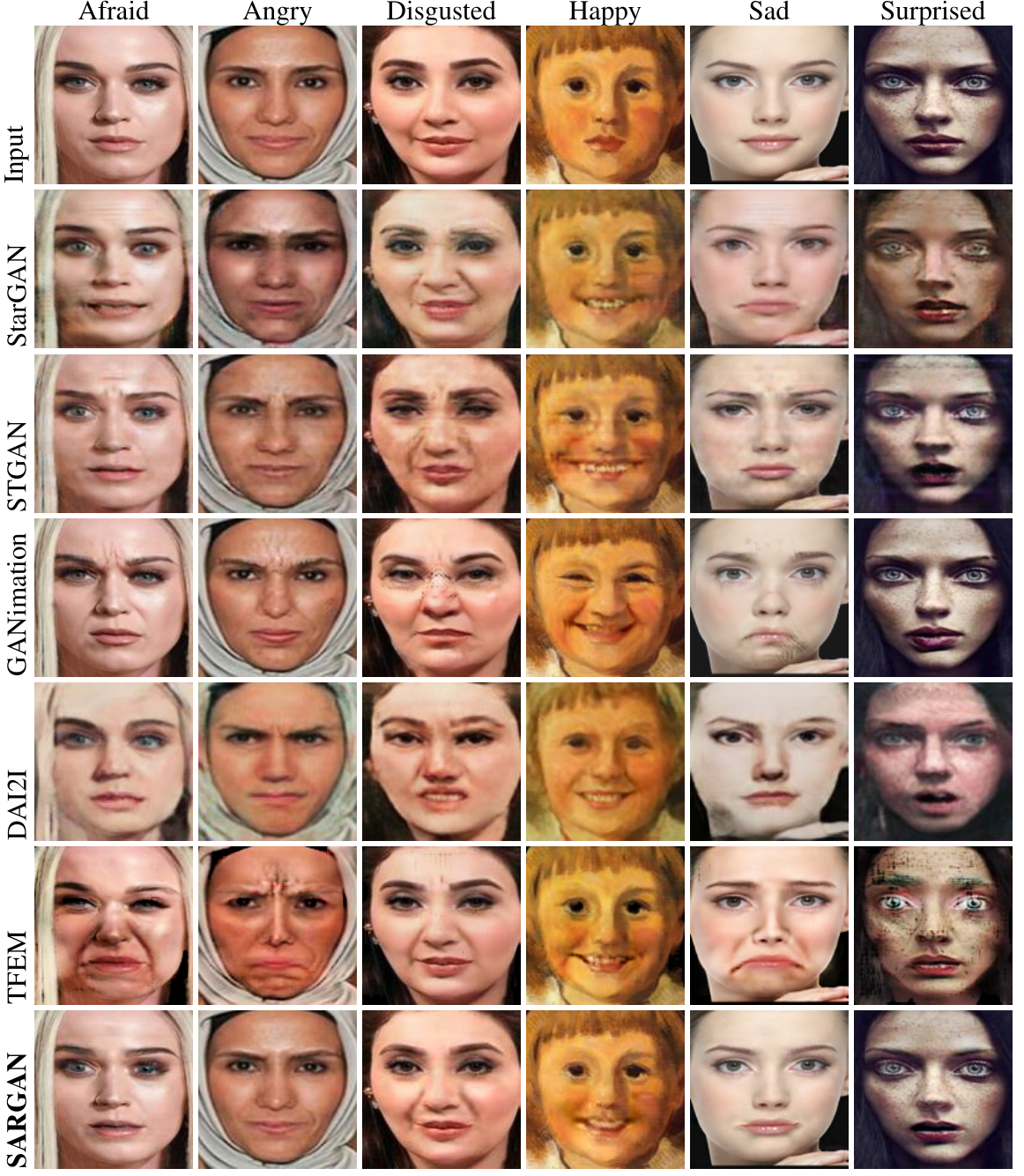}
     \caption{Comparison of facial expression manipulation results obtained by the proposed SARGAN and other state-of-the-art methods on out-of-distribution images. StarGAN and STGAN fail to recover the true input image facial and color details while GANimation introduces noticeable artifacts around expression-rich regions such as eyes, nose and mouth. DAI2I fails to preserve identity and facial details in synthesized images and introduces painting-like bias. TFEM tends to generate significant artifacts around the eyes, nose and mouth.
     The proposed method i) correctly induced the desired expressions, ii) did not introduce artifacts, and iii) preserved rich facial details such as iris reflections, makeup and skin pigmentation. } 
    \label{fig:comp-state-of-the-art}
\end{figure}

\begin{figure}[ht]
    \centering
    \includegraphics[width=\linewidth]{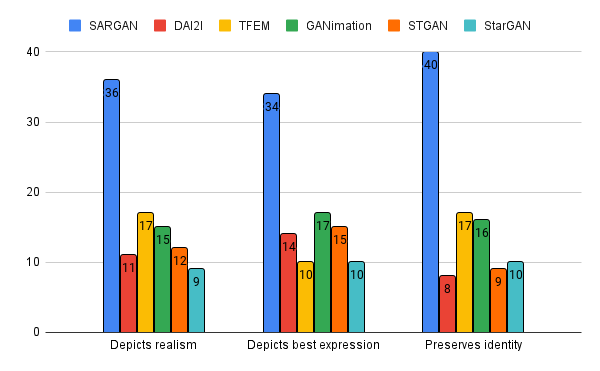}
    \caption{Results of the user study (numbers shown are percentages, higher is better).}
    \label{fig:user_study}
\end{figure}

\subsection{Evaluation Metrics}
To evaluate the visual quality of manipulated images  quantitatively, we use Average content distance (ACD), Face verification scores (FVS), Frechet inception distance (FID) and User study.\\
\textbf{Average content distance} 
calculates the $\ell_2$ distance between features of input and manipulated images extracted by openFace \footnote{\url{https://github.com/ageitgey/face_recognition}}. It is a deep model trained for face recognition task. \\
\textbf{Face verification score} 
computes the similarity between input and manipulated
images using Face++ tool \footnote{\url{https://www.faceplusplus.com/face-comparing/}}. \\
\textbf{Frechet inception distance}
measures the distance between feature vectors of real and synthesized faces \cite{heusel2017gans_fid}. \\
\textbf{User study} 
We conducted a user study to assess the quality of our proposed method. We randomly picked six images from testing images (CFEE, KDEF and RaFD) and twelve images from out-of-distribution images. Given one neutral input image, we have asked 35 human participants to select the best-manipulated image in an answer to these questions: i) Which image depicts realism? ii) Which image depicts best expression? and iii) Which image preserves identity? The options were six randomly shuffled manipulated images from StarGAN, STGAN, GANimation, DAI2I, TFEM as well as SARGAN. Sample image from user study is shown in Fig. \ref{fig:user_study}

\begin{figure*}[ht]
    \centering
    \includegraphics[width=.9\linewidth]{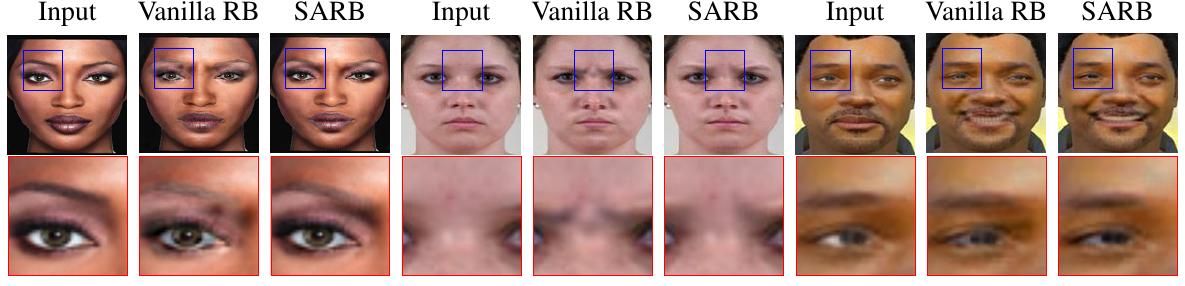}
    \caption{Results of SARGAN with vanilla residual block (Vanilla RB) and spatial attention-based residual block (SARB). SARB produces sharper and clearer images while preserving facial details the most than Vanilla RB.}
    \label{fig:spatial-vs-vanilla}
\end{figure*}

\subsection{Qualitative Evaluation}
Fig. \ref{fig:indataset-results} shows the results of SARGAN on in-dataset images. These results show that SARGAN successfully induces target expression while retaining the facial details including skin tone, eye color and background details. 
Our proposed method can be used to manipulate the expression of out-of-distribution images as illustrated in Fig. \ref{fig:outofdataset-results}. Fig. \ref{fig:outofdataset-results} shows that SARGAN introduces desired expression on a variety of images including painting (Panel 2, rows 1-4), statues (Panel 1, row 5 and Panel 3, row 4), avatar (Panel 1, row 4 and Panel 3, row 5), low-resolution images (Panel 1, row 3), occluded image by hand (Panel 1, row 2), occluded image by external objects (Panel 3, rows 2 and 3) and image with different light shades (Panel 2, row 5) while preserving rich facial details. 
Comparison of SARGAN with five state-of-the-art approaches \cite{pumarola2018ganimation} is presented in Fig. \ref{fig:comp-state-of-the-art}. We observe that GANimation \cite{pumarola2018ganimation} preserves the rich input image details due to attention mask but fails to induce the desired expression. It also introduces noticeable artifacts. On the contrary, StarGAN \cite{choi-2017} introduces the desired expression. However, it can be seen that its results lose rich facial details besides introducing artifacts. STGAN tends to loose the input image facial and color details and it also introduces artifacts around expression-rich regions such eyes, nose and mouth. TFEM fails to synthesize realistic afraid, angry and surprised expressions. DAI2I emulates oil painting style in the synthesized images, due to being trained on the oil painting dataset, and fails to preserve the identity details of input images. Compared to these existing approaches, SARGAN successfully induces the desired expression while keeping intact rich details of input faces. \revision{Our method works on these images because i) it was successful in locating human-like features, ii) then focusing on them via spatial attention, iii) then changing them to induce the required expression, and iv) finally transferring overall facial details via the residuals mechanism. This allows the model to induce realistic expressions on a variety of unseen images.}

\subsection{Quantitative Evaluation}
It can be seen in Table \ref{tab:qunatitative_evaluation} that SARGAN achieves the lowest Frechet inception distance (FID) among all other methods. The mean and standard deviation of face verification scores (FVS) and average content distance (ACD) on testing images show that our proposed method outperformed all other state-of-the-art methods in terms of identity preservation. As shown in Fig. \ref{fig:user_study}, SARGAN obtains the majority of votes in all three questions.

\begin{table}[t]
    \centering
    \begin{tabular}{|c|c|c|c|} \hline
    Method & ACD $\downarrow$ & FVS $\uparrow$ & FID $\downarrow$  \\ \hline
    SARGAN & \textbf{0.306} & \textbf{94.19} $\pm$ \textbf{1.11} &  \textbf{53.95}  \\ \hline
    DAI2I \cite{chen2020domain} & 0.485 & 84.64 $\pm$ 4.45 & 89.81 \\ \hline
    TFEM \cite{ling2020toward} & 0.442 & 85.31 $\pm$ 4.74  & 72.12 \\ \hline
    GANimation \cite{pumarola2018ganimation} & 0.389 & 87.98 $\pm$ 8.67 & 55.30 \\ \hline
    STGAN \cite{liu2019stgan} & 0.372 & 92.05 $\pm$ 2.81 & 72.46 \\ \hline
    StarGAN \cite{choi-2017} & 0.566 & 91.71 $\pm$ 1.95 & 92.46 \\ \hline
    \end{tabular}
    \caption{Quantitative results of proposed and state-of-the-art expression manipulation methods.}
    \label{tab:qunatitative_evaluation}
\end{table}

\begin{figure}[t]
    \centering
    \includegraphics[width=.95\linewidth]{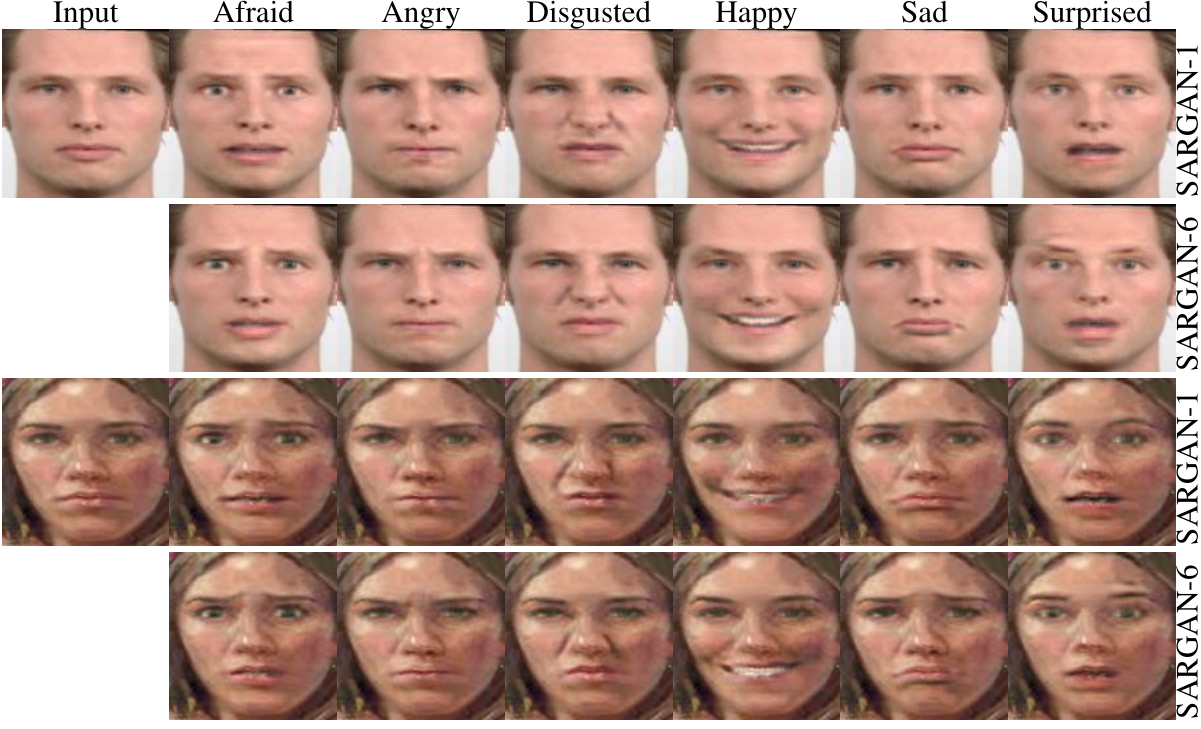}
    \caption{Comparison of SARGAN results obtained using one and six spatial attention residual blocks. Using six blocks does not seem to improve expression quality significantly. Please zoom in for more details.}
    \label{fig:six-vs-one-residual-block}
\end{figure}

\begin{figure}[t]
    \centering
    \includegraphics[width=.95\linewidth]{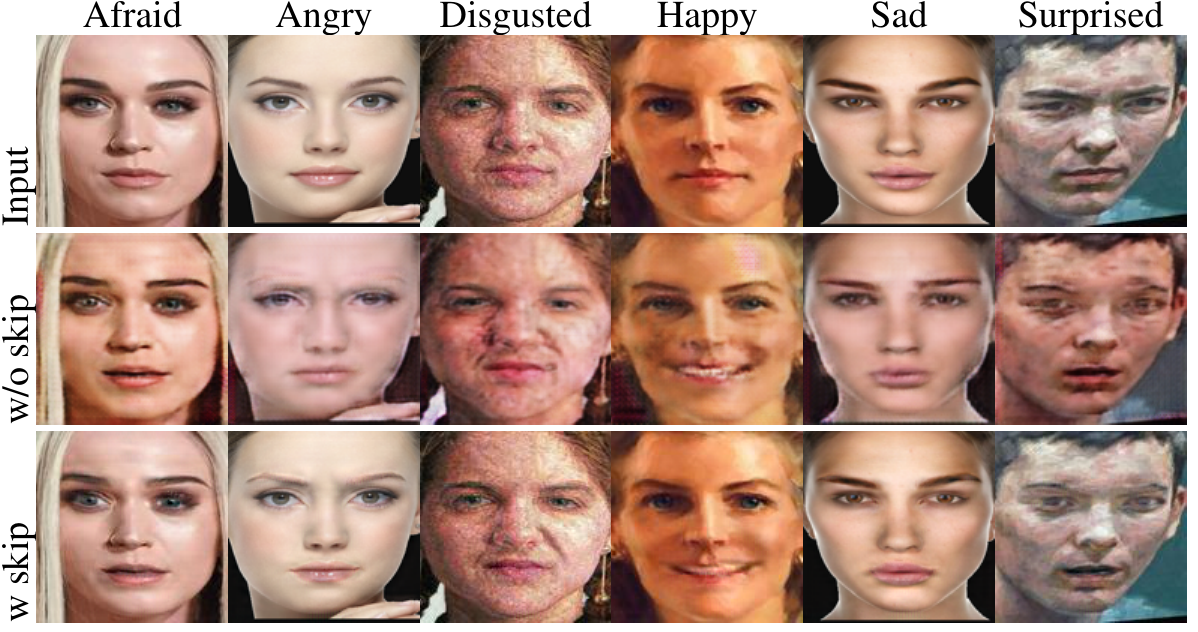}
    \caption{Ablation study of SARGAN over out-of-distribution images. The inclusion of residual connection helps to generate sharper expressions with fewer artifacts and rich input details. Most importantly, it helps to retain the overall color distribution of the input image. Please zoom in for more details.}
    \label{fig:with-vs-without-skip-connection}
\end{figure}

\begin{figure*}[ht]
    \centering
    \includegraphics[width=.8\linewidth]{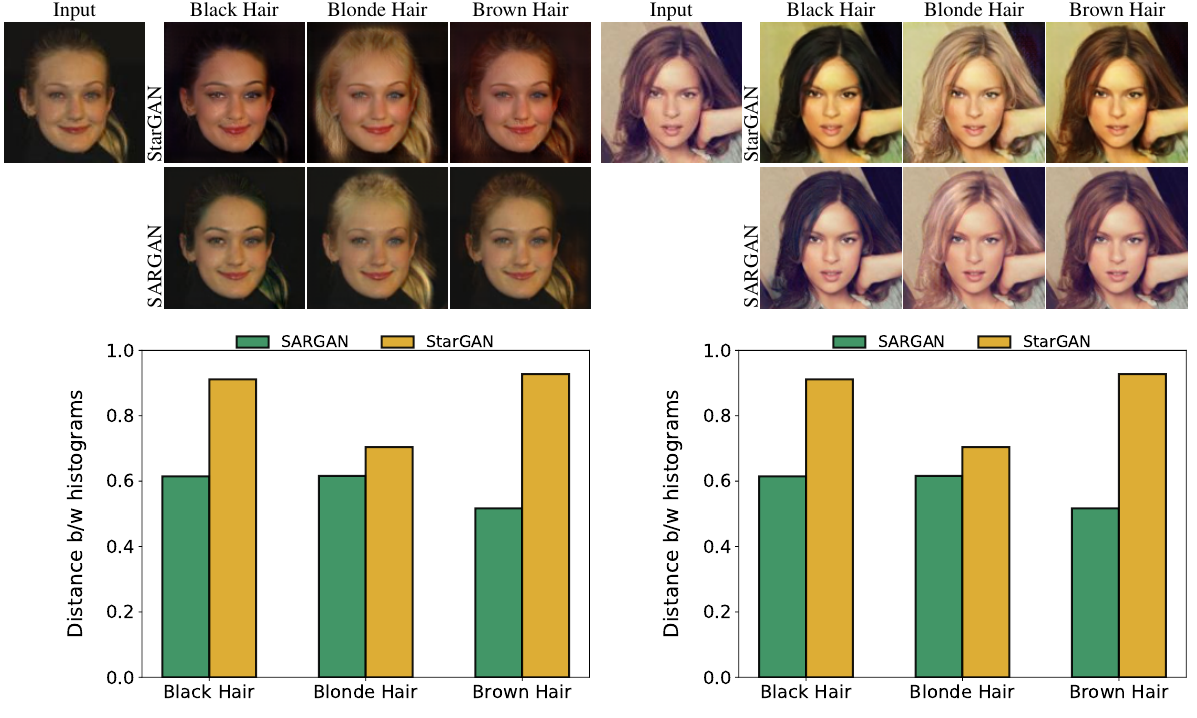}
    \caption{Comparison of facial attribute manipulation results obtained by the StarGAN model and the proposed SARGAN model. \textbf{Top}: StarGAN fails to recover the true colors and facial details of the input image such as eye and skin colors. SARGAN preserves the color and facial details of the input image while manipulating facial attributes. 
    \textbf{Bottom}: Histogram distances between synthesized outputs and input image. Compared to StargGAN, SARGAN better preserves the color distribution of input images as evidenced by the lower histogram distances.
    }
    \label{fig:attr_manipulation}
\end{figure*}

\begin{figure}[h]
    \centering
    \includegraphics[width=.95\linewidth]{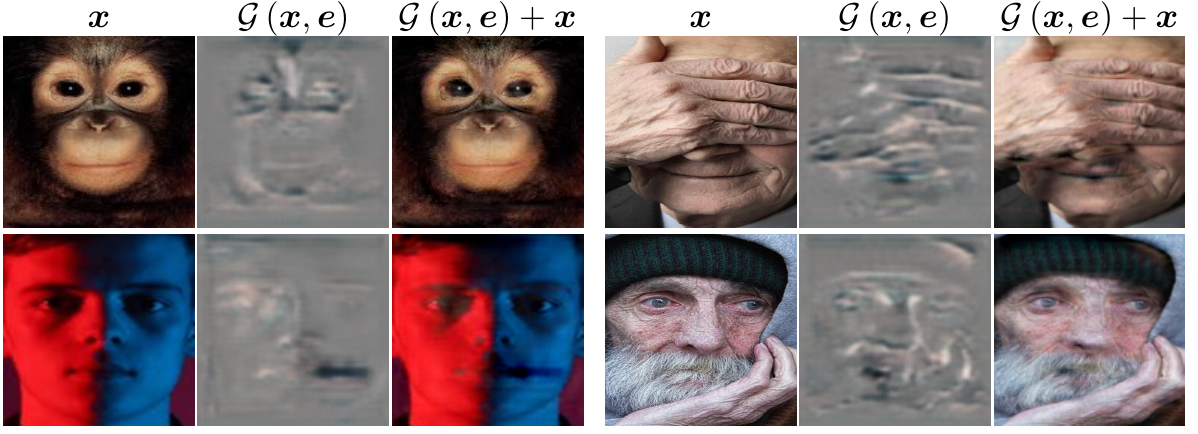}
    \caption{Failure cases. \textbf{Row 1}: SARGAN fails to introduce the surprised expression on a non-human face and the self-occluded face of an old man. The model did not yield correct surprised residuals for these images. \textbf{Row 2, Left}: SARGAN is unable to locate the correct facial features on the multi-colored face to change its expression from neutral to surprised. Residual image shows that the model tends to yield an open mouth in the wrong place. \textbf{Row 2, Right}: SARGAN tends to generate surprised expression as can be seen in the residual image. However, because of beard hair, the expression is not as evident.}
    \label{fig:failure_cases}
\end{figure}

\begin{figure*}[t]
    \centering
    \includegraphics[width=.85\linewidth]{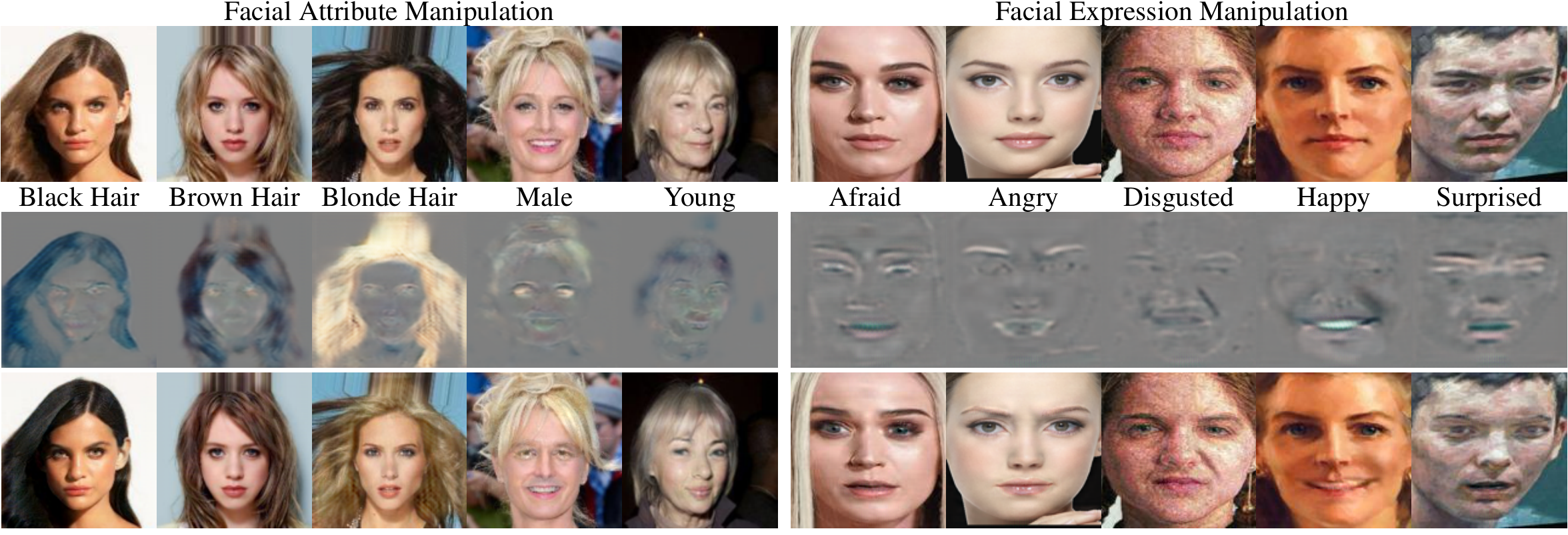}
    \caption{\revision{Illustration of the input image $\bm{x}$ (first row), residual $\mathcal{G}(\bm{x}, \bm{e})$ (second row) and final output $\mathcal{G}(\bm{x}, \bm{e}) + \bm{x}$ (third row) produced by SARGAN. Final residual connection helps to learn only expression-related details. The residuals of attribute manipulation are dense. While the residuals of expression manipulation are sparse. This indicates that expression manipulation causes localized changes in specific regions of images.}}
    \label{fig:residuals_of_fem_and_fam}
\end{figure*}

\subsection{Ablation Study}
To validate the effectiveness of each component of the proposed SARGAN, we pose three questions:
\vspace{0.2in}

\textit{\textbf{Why spatial attention based residual block instead of vanilla residual block?}} To validate the effectiveness of spatial attention-based residual block instead of vanilla residual block, we trained SARGAN with vanilla residual block. Fig. \ref{fig:spatial-vs-vanilla} shows that spatial attention-based SARGAN produces sharper and realistic expressions in addition to suppressing artifacts. 

\vspace{0.2in}

\textit{\textbf{Are six residual blocks effective for face-to-face translation?}} Image-to-image translation models such as CycleGAN \cite{zhu-2017}, StarGAN \cite{choi-2017}, RelGAN \cite{wu2019relgan} and GANimation \cite{pumarola2018ganimation} use six vanilla residual blocks in the "bottleneck" of their generator network for $128 \times 128$ size images. A key question arises here: \textit{why should we use six residual blocks?} We find that CycleGAN first employed six residual blocks for image-to-image translation problem inspired by Johnsons et al. \cite{johnson-2016perceptual}. CycleGAN has shown reasonable results on image-to-image translation problem. StarGAN, RelGAN and GANimation adapted the same architecture for face-to-face translation. We find that six residual blocks are not necessarily required for facial expression manipulation. This is intuitively demonstrated in Fig. \ref{fig:six-vs-one-residual-block}, where we train SARGAN with one and six residual blocks respectively. We denote models with one and six residual blocks as SARGAN-1 and SARGAN-6. It can be observed that the quality of results did not justify the six-fold increase in the number of parameters. It seems one spatial attention residual block might be sufficient for facial expression manipulation.

\vspace{0.2em}
\textit{\textbf{Why should we train SARGAN's generator with residual connection?}}
To demonstrate that the residual connection helps in retaining rich facial details, we trained our proposed method with and without a residual connection. Fig. \ref{fig:with-vs-without-skip-connection} shows that the residual connection helps to obtain photo-realistic expressions on out-of-distribution images which not only induces the desired expression but also perfectly preserves the rich facial details of input faces. \revision{The residual connection is indispensable for facial expression synthesis which involves transferring most of the color distribution of the input image while making sparse, local changes to affect a new expression. In contrast, the residual connection is not as important for general image-to-image translation and general facial attribute manipulation tasks that can require changing most of the color distribution of the input image.}

\subsection{Shortcomings}
Despite the fact that training SARGAN with residual connection helps in recovering facial and overall color details of the input image, it also reduces the expressiveness of expressions. Fig. \ref{fig:with-vs-without-skip-connection} shows that expressions without skip-connection are much more intense than those with skip-connection. To solve this problem, it may be possible to gate the skip connection using attention mechanism. Additionally, we illustrated some failure cases of SARGAN in Fig. \ref{fig:failure_cases}. It shows that the proposed method fail to elicit expression on animal faces (Fig. \ref{fig:failure_cases}, Row1 and Panel 1). Further, It is  incapable of inducing convincing expressions on a multi-colored face (Fig. \ref{fig:failure_cases}, Row2 and Panel 1) and an obscured faces (Fig. \ref{fig:failure_cases}, Panel 2).

\subsection{Is SARGAN effective for facial attribute manipulation?}
We trained the proposed SARGAN model on CelebA dataset to assess the performance of SARGAN on facial attribute manipulation. For comparison, StarGAN results were obtained via pre-trained model provided by their authors \cite{choi-2017}. The results in Fig. \ref{fig:attr_manipulation} demonstrate that StarGAN is unable to recover the true input image overall color as well as facial details such as eye and skin colors. In contrast, SARGAN recovers true input image colors and facial details. To confirm our analysis, we compute the distance between histograms of input and StarGAN-synthesized images and input and SARGAN-synthesized images, respectively. Fig. \ref{fig:attr_manipulation} shows that SARGAN results are closer to the input images compared to StarGAN. Further, as shown in Fig. \ref{fig:attr_manipulation}, Column 3, StarGAN modifies the input face's hairstyle in addition to the hair color, whereas SARGAN only modifies the hair color. These findings support our claim that SARGAN with SARB and skip connection is generally beneficial for facial image manipulation tasks. 

\subsection{Discussion}
In this subsection, we discuss the premise behind the inclusion of an end-to-end skip connection and SARB in facial manipulation.

\textbf{Why should an end-to-end skip connection benefit facial manipulation?}

Image-to-image translation (I2I) and Facial manipulation (FM) are two different problems. Converting an image from one representation of a given scene to another is called I2I translation. In contrast, FM takes a face image of the person with any expression or attribute and synthesizes a new face image of the \textit{same} person but with a different expression or attribute. I2I models may change the color or style of the input image, such as converting a grayscale image to color, transferring seasons, or transferring a painting-like style to an input image. An end-to-end skip connection that transfers input image details directly into the output can be counter-productive in this scenario. FM models, on the other hand,  should not change the color or style of the input image. They are intended to change only the local attribute or expression of the input image. Therefore, the proposed end-to-end skip connection in FM models directly copies the color and facial details from the input image, leaving the SARGAN generator to focus on expression or attribute synthesis only. \revision{Fig. \ref{fig:residuals_of_fem_and_fam} shows that the  residuals of both expression and attribute manipulation support our assertion.}

\textbf{Why should SARB benefit facial manipulation?}
Some facial features, such as the eyebrows, eyes, nose, and mouth, are more important in facial manipulation than others. In order to focus more on these informative facial features and suppressing uninformative ones, the spatial attention mechanism within a residual block helps in synthesizing sharper and clearer images in addition to suppressing artifacts.

\section{Conclusion}
\label{conclusion}
We revisit encoder-decoder architecture in the generator of state-of-the-art facial manipulation models and observe that these models have come up short in preserving facial details. We then propose a novel framework called SARGAN to address this issue. In SARGAN, we propose to extend vanilla residual block with spatial attention mechanism. We utilize a symmetric encoder-decoder network to attend facial features at multiple scales and we propose to train the SARGAN generator  with residual  connection, to preserve facial details. Spatial attention block and residual connection make the training of facial expression manipulation more effective and efficient. Experiments substantiate the effectiveness of our proposed method over existing state-of-the-art solutions. Our work has shown that a GAN-based model can be trained using smaller datasets. Existing facial expression manipulation models fail to generalize well on out-of-distribution images. In contrast, SARGAN demonstrates better generalization on out-of-distribution images including human photographs, portraits, avatars and statues.

\bibliographystyle{IEEEtran}
\bibliography{egbib}

\end{document}